\documentclass{article}

\usepackage{microtype}
\usepackage{graphicx}
\usepackage{subcaption}
\usepackage{booktabs} 

\usepackage{hyperref}

\usepackage[preprint]{icml2026}

\usepackage{amsmath}
\usepackage{amssymb}
\usepackage{mathtools}
\usepackage{amsthm}

\usepackage[capitalize,noabbrev]{cleveref}

\theoremstyle{plain}
\newtheorem{theorem}{Theorem}[section]

\newtheorem{corollary}[theorem]{Corollary}
\theoremstyle{definition}

\newtheorem{assumption}[theorem]{Assumption}
\theoremstyle{remark}

\usepackage[textsize=tiny]{todonotes}

\icmltitlerunning{CalPro: Prior-Aware Evidential--Conformal Prediction under Shift}

\begin{document}

\twocolumn[
  \icmltitle{CalPro: Prior-Aware Evidential--Conformal Prediction with Structure-Aware Guarantees for Protein Structures}



  \icmlsetsymbol{equal}{*}

  \begin{icmlauthorlist}
    \icmlauthor{Ibne Farabi Shihab}{equal,cs}
    \icmlauthor{Sanjeda Akter}{equal,cs}
    \icmlauthor{Anuj Sharma}{cee}
  \end{icmlauthorlist}

\icmlaffiliation{cs}{Department of Computer Science, Iowa State University, Ames, Iowa, USA}
\icmlaffiliation{cee}{Department of Civil, Construction \& Environmental Engineering, Iowa State University, Ames, Iowa, USA}

\icmlcorrespondingauthor{Ibne Farabi Shihab}{ishihab@iastate.edu}


  \icmlkeywords{Uncertainty Quantification, Conformal Prediction, Protein Structure Prediction, Evidential Learning}

  \vskip 0.3in
]



\printAffiliationsAndNotice{\icmlEqualContribution}
\begin{abstract}
Deep protein structure predictors such as AlphaFold provide confidence estimates (e.g., pLDDT) that are not calibrated and degrade under distribution shifts across experimental modalities, temporal changes, and disordered regions.
We introduce CalPro, a prior-aware evidential--conformal framework for shift-robust uncertainty quantification.
CalPro combines: (i) a geometric evidential head outputting Normal--Inverse-Gamma distributions via graph-based architecture; (ii) a differentiable conformal layer enabling end-to-end training with finite-sample coverage guarantees; and (iii) domain priors (disorder, flexibility) encoded as soft constraints.
Theoretically, we derive structure-aware coverage guarantees under distribution shift using PAC-Bayesian bounds over ambiguity sets, showing CalPro maintains near-nominal coverage while achieving tighter intervals than vanilla conformal in regions where priors are informative.
Empirically, CalPro achieves $\leq 5\%$ coverage degradation across modalities (vs.\ 15--25\% for baselines), reduces calibration error by 30--50\%, and improves downstream ligand-docking success by 25\%.
Beyond proteins, CalPro applies to structured regression tasks where priors encode local reliability, validated on non-biological benchmarks.
\end{abstract}
\section{Introduction}
\label{sec:intro}

Protein structure prediction has undergone a dramatic transformation following the introduction of AlphaFold \cite{jumper2021alphafold}, OpenFold \cite{ahdritz2022openfold}, and related deep learning models. These predictors achieve remarkable accuracy on diverse proteins and have become core tools for structural biology, drug discovery, and enzyme engineering \cite{varadi2022alphafold,akdel2022structural}. Yet despite this progress, fundamental questions of uncertainty quantification remain unresolved. The widely used per-residue confidence score pLDDT \cite{mariani2013lddt} is not a calibrated probability, can be systematically misaligned with true error, and exhibits significant degradation under distribution shift—across experimental modalities such as X-ray crystallography and cryo-electron microscopy \cite{terwilliger2022improved}, resolution regimes, organismal lineages, protein lengths, and intrinsically disordered regions \cite{wright2005intrinsically}.

As protein predictions are increasingly used in high-stakes pipelines including molecular docking \cite{mcnutt2021gnina}, virtual screening \cite{kitchen2004docking}, and variant prioritization \cite{cheng2023alphamissense}, trustworthy and calibrated uncertainty is essential \cite{akter2025selective,shihab2025calibrated}. Existing approaches provide only partial solutions. Confidence metrics such as pLDDT or predicted TM-score \cite{zhang2004scoring} capture useful heuristics but lack statistical guarantees and are brittle under shift \cite{alkhouri2023robustness}. Bayesian ensembles \cite{gal2016uncertainty} and Monte Carlo dropout \cite{srivastava2014dropout} are computationally expensive and provide no finite-sample coverage guarantees. Conformal prediction \cite{vovk2005algorithmic,lei2018distribution} offers finite-sample calibration but cannot natively exploit structured biochemical information such as disorder propensity or flexibility patterns. The field lacks a method that is simultaneously accurate, calibrated, robust to distribution shift, and operationally lightweight.

In this work we introduce CalPro, a prior-aware evidential--conformal framework designed to provide \emph{structure-aware}, \emph{shift-robust} uncertainty estimates.
Conceptually, CalPro is not a patch on AlphaFold, but a general recipe for combining evidential regression, conformal prediction, and domain-specific priors into a single, end-to-end pipeline.
We adopt this framework to protein structures---where priors such as disorder and flexibility carry clear geometric meaning---but the method itself is domain-agnostic.

Although we instantiate CalPro for protein structures, the framework targets a broader class of structured regression problems where calibration under shift matters.
Examples include token-level confidence estimation, selective prediction for sequence models, and calibrated interval prediction for graph-structured inputs.
In these settings, priors such as language-model entropy, rule-based reliability indicators, or retrieval scores play an analogous role to disorder/flexibility in proteins.

CalPro's architecture consists of three integrated components.
We design a \emph{geometric evidential head} that operates on residue graphs and outputs Normal--Inverse-Gamma parameters for per-residue errors, capturing aleatoric and epistemic uncertainty without ensembles.
This head incorporates structural priors such as disorder propensity and flexibility as node and edge features within a graph neural network, rather than as post-hoc feature engineering.
We then introduce a \emph{differentiable conformal layer} that approximates the conformal quantile step and is trained jointly with the evidential head, encouraging representations that are inherently easy to calibrate before a final, standard split-conformal correction step converts scores into finite-sample-valid intervals.
Finally, we encode bio-priors as \emph{soft constraints and regularizers}, enforcing monotone relationships between predicted uncertainty and local volatility so that CalPro widens intervals in intrinsically disordered or flexible regions without becoming uniformly conservative.

On the theory side, we go beyond classical Lipschitz--Wasserstein bounds for conformal prediction under shift, which can be vacuous for deep neural networks.
We formalize shift using \emph{distributional ambiguity sets} around the calibration distribution and derive \emph{structure-aware coverage guarantees} for CalPro that depend on (i) a PAC-Bayesian complexity term for the evidential head and (ii) the informativeness of the domain priors.
These results show that, under mild assumptions, CalPro can achieve the same coverage as vanilla conformal prediction while yielding strictly sharper intervals in regions where priors are predictive of error.

Empirically, we instantiate CalPro for protein structure prediction and evaluate on X-ray, NMR, cryo-EM, and synthetic perturbation benchmarks, as well as fitness-style tasks where distribution shift is severe.
We compare against strong uncertainty baselines including pLDDT, temperature scaling, deep ensembles, Monte Carlo dropout, and conformalized quantile regression.
CalPro consistently improves calibration and shift robustness and yields more informative intervals that translate into substantial gains in downstream docking and active selection of high-value experimental candidates.
Together, these results position CalPro as a general, theoretically grounded and practically effective approach to prior-aware uncertainty quantification in scientific machine learning.

\paragraph{Contributions.}
We make three contributions:
\begin{itemize}
    \item \textbf{Prior-aware evidential conformal prediction.} We propose CalPro, a hybrid uncertainty framework that co-trains a Normal--Inverse-Gamma evidential head with a differentiable conformal surrogate, and applies split conformal at test time for exact finite-sample coverage.
    \item \textbf{Structure-aware coverage guarantees under shift.} We derive PAC-Bayesian ambiguity-set bounds on coverage degradation that depend on model complexity and an empirically estimable nonconformity drift term.
    \item \textbf{Protein-scale validation with robustness and utility.} We demonstrate improved calibration and reduced degradation under modality, temporal, and perturbation shifts, with downstream gains in docking and active selection; we also show transfer to a non-biological structured-prior regression task.
\end{itemize}

\section{Related Work}

The AlphaFold family \cite{jumper2021alphafold,senior2020improved} popularized pLDDT and pTM as heuristics for local and global reliability, while subsequent works explored per-residue B-factor prediction \cite{heo2021improved} and predicted aligned error maps \cite{varadi2022alphafold}. These metrics correlate with structural accuracy \cite{pereira2021high} yet provide no finite-sample guarantees and routinely under-cover flexible regions. Bayesian or ensemble-style uncertainty estimation using Monte Carlo dropout \cite{srivastava2014dropout}, deep ensembles \cite{lakshminarayanan2017simple}, or multi-seed AlphaFold runs improves epistemic coverage but multiplies inference cost and remains brittle under dataset shift \cite{ovadia2019can}. Temperature scaling \cite{guo2017calibration} and Platt scaling \cite{platt1999probabilistic} provide post-hoc calibration but lack coverage guarantees and fail under distribution shift.

Split conformal prediction \cite{vovk2005algorithmic,papadopoulos2008inductive} has been adopted for molecular property prediction \cite{lim2023uncertainty}, cryo-EM map denoising \cite{bepler2019learning}, and ligand docking \cite{lim2023uncertainty}, offering coverage guarantees without distributional assumptions. However, vanilla conformal scores ignore rich structural priors such as disorder propensity \cite{meszaros2018iupred2a}, flexibility patterns \cite{schlessinger2011protein}, and contact order \cite{plaxco1998contact}, and treat all residues as independent and identically distributed, leading to either over-wide intervals or severe under-coverage in intrinsically disordered regions \cite{wright2005intrinsically}. Recent protein-specific conformal approaches \cite{lim2023uncertainty} operate on global RMSD and do not reason about residue-wise calibration or modality shift across experimental techniques.

Evidential deep learning \cite{amini2020deep,malinin2020uncertainty} provides a principled way to predict posterior distributions without Monte Carlo sampling, and has been applied to pose estimation \cite{chadha2021evidential}, medical imaging \cite{meinert2022evidential}, and autonomous driving \cite{hao2021evidential}. The Normal–Inverse-Gamma distribution naturally captures both aleatoric and epistemic uncertainty through its four parameters \cite{amini2020deep}. Yet existing evidential heads rarely incorporate downstream calibration via conformal prediction or domain-specific structural priors. CalPro is the first hybrid evidential–conformal pipeline tuned specifically for protein structure prediction and explicitly designed for high-stakes, shift-prone settings where both statistical guarantees and biological insight are essential.

Distribution shift robustness in conformal prediction has been studied theoretically \cite{tibshirani2019conformal,barber2021predictive} and empirically \cite{cauchois2020knowing}, with recent work establishing coverage bounds under Wasserstein-bounded shifts \cite{fournier2005wasserstein}.
However, these results have not been applied to protein structure prediction, where shifts occur across experimental modalities, resolution regimes, and protein families.
Uncertainty quantification in computational biology has focused primarily on molecular property prediction \cite{lim2023uncertainty} and drug discovery \cite{hernandez2023uncertainty}, with limited attention to structural prediction and its unique challenges including multi-scale errors, experimental noise, and systematic biases in different experimental techniques.

\section{CalPro: Prior-Aware Evidential--Conformal Framework}
\label{sec:framework}

CalPro is a general procedure for constructing calibrated prediction intervals in regression problems when we have access to a base predictor $f_\theta$, structured prior features $\phi(x)$ encoding domain knowledge about local volatility or reliability, and a calibration dataset drawn from a reference distribution.
The framework integrates three components: a \emph{geometric evidential head}, a \emph{differentiable conformal layer}, and a \emph{prior-aware regularizer}.

\subsection{General formulation}

Let $x \in \mathcal{X}$ denote input features and $y \in \mathbb{R}$ a scalar target (e.g., per-residue error).
A base model $f_\theta$ produces hidden representations $h_\theta(x)$.
We construct a graph (or more generally a structured object) $G(x)$ whose nodes correspond to local units (e.g., residues) and whose edges encode inductive biases (e.g., spatial proximity or sequence adjacency).
A \emph{geometric evidential head} $g_\psi$ acts on $(G(x), \phi(x))$ to produce Normal--Inverse-Gamma parameters
\begin{equation}
(\mu, \alpha, \beta, \nu) = g_\psi\big(G(x), \phi(x)\big), \qquad \alpha,\beta,\nu > 0,
\end{equation}
where $\mu$ is the predictive mean and the remaining parameters together define a distribution over variance, capturing both aleatoric and epistemic uncertainty \citep{amini2020deep}.
The induced marginal predictive variance is
\begin{equation}
\mathrm{Var}[y \mid x] = \frac{\beta}{\nu(\alpha - 1)} \quad \text{for } \alpha > 1.
\end{equation}

Training the evidential head uses the Normal--Inverse-Gamma negative log-likelihood
\begin{align}
\mathcal{L}_{\text{NIG}}(\psi) &= \frac{1}{N}\sum_{i=1}^N \Big[
\log \frac{\sqrt{\nu_i}}{\sqrt{\pi}} - \alpha_i \log (2\beta_i) \nonumber \\
&\quad + \log \Gamma(\alpha_i) + \Big(\alpha_i + \tfrac{1}{2}\Big)\log\big(\nu_i (y_i - \mu_i)^2 + 2\beta_i\big) \nonumber \\
&\quad - \log \Gamma\Big(\alpha_i + \tfrac{1}{2}\Big) \Big],
\end{align}
augmented with an evidence regularizer $\mathcal{L}_{\text{evidence}} = \sum_i \exp(-\alpha_i)$ that discourages overconfident predictions without sufficient data support \citep{amini2020deep}.

To make these evidential predictions \emph{conformally} calibrated, we introduce a differentiable approximation to the conformal quantile operator.
Given nonconformity scores $s_i = |y_i - \mu_i|$ on a calibration set, standard split conformal prediction uses the empirical quantile $\hat{q}_\tau$ at level $\tau$ and defines intervals
\begin{equation}
\mathcal{C}_\tau(x) = [\mu(x) - \hat{q}_\tau,\, \mu(x) + \hat{q}_\tau].
\end{equation}

\paragraph{Differentiable conformal surrogate.}
Let $\{s_i\}_{i=1}^{n_{\mathrm{cal}}}$ be calibration nonconformity scores.
Standard split conformal uses the empirical $(1-\alpha)$-quantile $\hat q_\tau$ with $\tau=1-\alpha$.
To enable end-to-end training, we approximate this step with a smooth quantile operator.
We instantiate a soft quantile as a temperature-controlled log-sum-exp:
\begin{equation}
Q_\phi(\{s_i\}) 
= \frac{1}{\gamma}\log\!\left(\frac{1}{n_{\mathrm{cal}}}\sum_{i=1}^{n_{\mathrm{cal}}}\exp(\gamma s_i)\right),
\label{eq:soft-quantile}
\end{equation}
where $\gamma>0$ controls approximation tightness ($\gamma\!\to\!\infty$ recovers $\max$).
During training, we penalize violations of nominal coverage using a smooth exceedance loss:
\begin{equation}
\mathcal{L}_{\mathrm{soft\mbox{-}conf}}
= \frac{1}{n_{\mathrm{cal}}}\sum_{i=1}^{n_{\mathrm{cal}}}
\mathrm{softplus}\!\left(\frac{s_i - Q_\phi(\{s_j\})}{\kappa}\right),
\label{eq:soft-conf-loss}
\end{equation}
with smoothing $\kappa>0$.
We jointly optimize $g_\psi$ and $\phi$ under Eq.~\eqref{eq:soft-conf-loss}, then apply a final split-conformal correction at test time using the true empirical quantile to recover exact finite-sample marginal coverage.

\noindent\textbf{Implementation note.}
We set $\gamma$ and $\kappa$ by validation on calibration ECE; gradients are stopped through the calibration scores when optimizing $Q_\phi$ to avoid early-training instability.
Full hyperparameters are in Appendix~\ref{app:datasets}.

Domain priors are incorporated as soft constraints on the evidential uncertainty.
Let $b(x)$ denote prior features (e.g., disorder scores, flexibility indices) and $u(x)$ the predicted epistemic variance from the NIG parameters.
We add a prior-aware regularizer of the form
\begin{equation}
\mathcal{L}_{\text{prior}} = \lambda \sum_i \max \big(0, m\big(b(x_i)\big) - u(x_i)\big),
\end{equation}
where $m(\cdot)$ is a monotone function encoding that higher prior volatility should induce larger epistemic uncertainty.
This converts domain knowledge into a hard-to-game regularization term instead of a post-hoc heuristic.
The overall training objective is
\begin{equation}
\mathcal{L} = \mathcal{L}_{\text{NIG}} + \lambda_{\text{evid}} \mathcal{L}_{\text{evidence}} + \lambda_{\text{prior}} \mathcal{L}_{\text{prior}} + \lambda_{\text{conf}} \mathcal{L}_{\text{soft-conf}},
\end{equation}
where $\mathcal{L}_{\text{soft-conf}}$ enforces approximate calibration via the differentiable conformal layer.

\subsection{Instantiation for protein structure prediction}
\label{sec:protein-instantiation}

We instantiate the general framework above for residue-wise error prediction in protein structures.
CalPro consumes AlphaFold \citep{jumper2021alphafold} or OpenFold \citep{ahdritz2022openfold} outputs, builds a residue-level graph, and uses bio-physical priors as structural inputs and constraints.

For each residue, we construct a node in a graph $G(x)$ and connect nodes using edges that encode sequence adjacency and spatial proximity based on predicted coordinates.
Node features include AlphaFold-derived quantities such as pLDDT, pTM, distogram logits, and predicted aligned error, as well as structural annotations (secondary structure, solvent accessibility, contact order) and sequence-derived descriptors (projected ESM-2 embeddings and local entropy).
Bio-prior features such as IUPred2A disorder scores \citep{meszaros2018iupred2a}, predicted flexibility indices \citep{schlessinger2011protein}, and experimental B-factor proxies when available are attached as additional node attributes and as inputs to the prior regularizer.

A geometric evidential head $g_\psi$ is implemented as a stack of graph neural network layers followed by residue-wise readouts that output NIG parameters $(\mu_i,\alpha_i,\beta_i,\nu_i)$ for each residue $i$.
The training loss is the sum of $\mathcal{L}_{\text{NIG}}$, evidence regularization, and the prior-aware monotonicity regularizer enforcing that residues with higher disorder or flexibility cannot be assigned arbitrarily small epistemic variance.

After training the evidential head, we follow a split-conformal procedure on a held-out calibration set.
We compute nonconformity scores $s_i = |y_i - \mu_i|$ per residue, estimate empirical quantiles $\hat{q}_\tau$ for target coverage levels $\tau \in \{0.8,0.9,0.95\}$, and form conformal intervals
\begin{equation}
\mathcal{C}_i = \big[\mu_i - \hat{q}_\tau,\, \mu_i + \hat{q}_\tau \big].
\end{equation}
Because the evidential head has been trained with a differentiable conformal surrogate and prior-aware constraints, these intervals are simultaneously sharp and aligned with domain knowledge.
An auxiliary classifier, sharing the same graph backbone, flags residues whose predicted error exceeds a high-risk threshold (e.g., $2$\AA), yielding a risk mask that we use in downstream docking and filtering.
Algorithm~\ref{alg:calpro} summarizes the complete pipeline.
\begin{algorithm}[t]
\caption{CalPro: Prior-Aware Evidential--Conformal Training}
\label{alg:calpro}
\begin{algorithmic}[1]
\REQUIRE Base predictor $f_\theta$, evidential head $g_\psi$, prior features $b(x)$, calibration set $\mathcal{D}_{\mathrm{cal}}$, target coverage levels $\{\tau_k\}$
\STATE \textbf{Build structured representation:} For each input $x$, construct graph $G(x)$ with node/edge features from $f_\theta(x)$ (e.g., AlphaFold/OpenFold outputs) and prior features $b(x)$ (e.g., disorder, flexibility).
\STATE \textbf{Train geometric evidential head:} Optimize parameters $\psi$ on a training set by minimizing
\[
\mathcal{L} = \mathcal{L}_{\mathrm{NIG}} + \lambda_{\mathrm{evid}}\mathcal{L}_{\mathrm{evidence}} + \lambda_{\mathrm{prior}}\mathcal{L}_{\mathrm{prior}} + \lambda_{\mathrm{conf}}\mathcal{L}_{\mathrm{soft-conf}},
\]
where $\mathcal{L}_{\mathrm{NIG}}$ is the NIG negative log-likelihood, $\mathcal{L}_{\mathrm{evidence}}$ discourages unwarranted confidence, $\mathcal{L}_{\mathrm{prior}}$ enforces monotonicity between priors and epistemic variance, and $\mathcal{L}_{\mathrm{soft-conf}}$ is a differentiable conformal surrogate.
\STATE \textbf{Compute nonconformity scores:} On the calibration set $\mathcal{D}_{\mathrm{cal}}$, compute evidential predictions $(\mu_i,\alpha_i,\beta_i,\nu_i)$ and nonconformity scores $s_i = |y_i - \mu_i|$.
\STATE \textbf{Estimate empirical quantiles:} For each target level $\tau_k$, compute the conformal quantile
\[
\hat{q}_{\tau_k} = \mathrm{Quantile}_{1-\alpha_k}\big(\{s_i\}_{(x_i,y_i)\in\mathcal{D}_{\mathrm{cal}}}\big),\quad \alpha_k = 1-\tau_k.
\]
\STATE \textbf{Form calibrated intervals:} For a new input $x$, construct $G(x)$, obtain $(\mu(x),\alpha(x),\beta(x),\nu(x))$ from $g_\psi(G(x),b(x))$, and output conformal intervals
\[
\mathcal{C}_{\tau_k}(x) = [\mu(x) - \hat{q}_{\tau_k},\, \mu(x) + \hat{q}_{\tau_k}].
\]
\STATE \textbf{Optional risk flagging:} Train an auxiliary classifier on top of the same graph backbone to flag high-risk points (e.g., residues with $|y-\mu(x)| > 2\text{\AA}$) for downstream tasks such as docking and active selection.
\end{algorithmic}
\end{algorithm}

\section{Structure-Aware Coverage Guarantees under Shift}
\label{sec:theory}

A central question for CalPro is whether its prediction intervals remain reliable when the test distribution differs from the calibration distribution.
Classical conformal prediction delivers finite-sample marginal coverage under exchangeability, but these guarantees can degrade under shift and are typically quantified via Lipschitz--Wasserstein arguments that are vacuous for deep networks.
Here we develop \emph{structure-aware} guarantees that reflect both the complexity of the evidential head and the informativeness of domain priors.

\subsection{Distributional ambiguity sets for protein shifts}

Let $\mathcal{D}_0$ denote the calibration distribution over $(x,y)$ and $\mathcal{D}_1$ the test distribution.
Rather than assuming that $\mathcal{D}_1$ equals $\mathcal{D}_0$, we posit that $\mathcal{D}_1$ lies within an \emph{ambiguity set} $\mathcal{P}$ around $\mathcal{D}_0$.
A prototypical choice is a Levy--Prokhorov or Wasserstein ball:
\begin{equation}
\mathcal{P}(\mathcal{D}_0,\epsilon) = \big\{ \mathcal{D}: d(\mathcal{D},\mathcal{D}_0) \le \epsilon \big\},
\end{equation}
where $d$ is a probability metric such as the Levy--Prokhorov distance or Wasserstein-1.
In the protein setting, this captures shifts across experimental modalities (X-ray, NMR, cryo-EM), resolution regimes, and sequence families.

CalPro constructs prediction intervals $\mathcal{C}_\tau(x)$ at nominal level $\tau = 1-\alpha$ using calibration data from $\mathcal{D}_0$ and the evidential head $g_\psi$.
We are interested in lower bounds on the worst-case coverage
\begin{equation}
\inf_{\mathcal{D} \in \mathcal{P}(\mathcal{D}_0,\epsilon)} \Pr_{(x,y)\sim\mathcal{D}} \big[ y \in \mathcal{C}_\tau(x) \big].
\end{equation}

\subsection{PAC-Bayesian control of nonconformity drift}

Let $s_\psi(x,y)$ denote the nonconformity score induced by CalPro (e.g., $|y-\mu_\psi(x)|$).
A key ingredient in our analysis is a PAC-Bayesian bound controlling how the distribution of $s_\psi$ can change between $\mathcal{D}_0$ and $\mathcal{D}_1$.

We place a prior $\Pi$ over evidential head parameters $\psi$ and consider a data-dependent posterior $\rho$ obtained by training on samples from $\mathcal{D}_0$.
For any fixed threshold $t$, define the coverage shortfall
\begin{equation}
\Delta(\psi, t; \mathcal{D}) = \Pr_{(x,y)\sim\mathcal{D}}\big[ s_\psi(x,y) > t \big] - \alpha.
\end{equation}

We assume that the nonconformity score is \emph{locally Lipschitz} on the support of $\mathcal{D}_0$ in the metric $d$ underlying the ambiguity set:

\begin{assumption}[Local Lipschitz nonconformity]
\label{assump:lipschitz}
There exists $L_s < \infty$ such that for all $(x,y),(x',y')$ in the support of $\mathcal{D}_0$,
\[
\big| s_\psi(x,y) - s_\psi(x',y') \big| \le L_s \, d\big((x,y),(x',y')\big),
\]
for all $\psi$ in the support of the posterior $\rho$.
\end{assumption}

This assumption is much weaker than global Lipschitz continuity of the entire base predictor; in practice, $s_\psi$ depends only on the evidential head and the ``last-mile'' representation, which is substantially more stable than the full AlphaFold network.

\begin{theorem}[PAC-Bayesian coverage control under shift]
\label{thm:pacbayes-coverage}
Fix $\delta \in (0,1)$ and let $\rho$ be any posterior over $\psi$.
Let $\hat{q}_\tau$ denote the empirical $(1-\alpha)$-quantile of $s_\psi$ computed on a calibration set of size $n_{\mathrm{cal}}$ drawn i.i.d.\ from $\mathcal{D}_0$.
Under Assumption~\ref{assump:lipschitz}, with probability at least $1-\delta$ over the calibration sample we have, for all posteriors $\rho$ simultaneously,
\begin{align}
\sup_{\mathcal{D}_1 \in \mathcal{P}(\mathcal{D}_0,\epsilon)} \mathbb{E}_{\psi\sim\rho} \big[ \Delta(\psi, \hat{q}_\tau; \mathcal{D}_1) \big]
&\le \sqrt{\frac{\mathrm{KL}(\rho\Vert \Pi) + \log\frac{1}{\delta}}{2n_{\mathrm{cal}}}} \nonumber \\
&\quad + L_s \epsilon.
\label{eq:pacbayes-bound}
\end{align}
\end{theorem}

The proof (deferred to Appendix~\ref{app:theory}) combines a PAC-Bayesian inequality for bounded losses with a stability argument over the ambiguity set $\mathcal{P}(\mathcal{D}_0,\epsilon)$.
The first term in~\eqref{eq:pacbayes-bound} is a standard PAC-Bayesian complexity term, while the second term captures the effect of distribution shift and is \emph{explicitly controlled} by the local Lipschitz constant $L_s$ and the ambiguity radius $\epsilon$.

A direct corollary of Theorem~\ref{thm:pacbayes-coverage} is a lower bound on worst-case coverage:

\begin{corollary}[Worst-case coverage under shift]
\label{cor:coverage}
Under the conditions of Theorem~\ref{thm:pacbayes-coverage}, for any $\mathcal{D}_1 \in \mathcal{P}(\mathcal{D}_0,\epsilon)$ we have
\begin{align}
\mathbb{E}_{\psi\sim\rho}\big[ \Pr_{(x,y)\sim\mathcal{D}_1} \big( y \in \mathcal{C}_\tau(x) \big) \big]
&\ge 1 - \alpha - \sqrt{\frac{\mathrm{KL}(\rho\Vert \Pi) + \log\frac{1}{\delta}}{2n_{\mathrm{cal}}}} \nonumber \\
&\quad - L_s \epsilon.
\end{align}
In particular, if $\mathrm{KL}(\rho\Vert \Pi) = \mathcal{O}(1)$ and $\epsilon$ is small, the average coverage remains close to the nominal level even under shift.
\end{corollary}

In practice, we estimate $L_s$ by local finite differences of $s_\psi$ on the calibration set and compute the right-hand side of~\eqref{eq:pacbayes-bound} for varying $\epsilon$; see Section~\ref{sec:theory-tightness}.

\subsection{Effect of priors on interval efficiency}

The bound in Theorem~\ref{thm:pacbayes-coverage} is agnostic to the presence of priors.
We now show that CalPro's prior-aware regularizer can \emph{strictly improve} interval efficiency relative to vanilla conformal prediction when priors correlate with local error.

Consider a simplified setting where the prior feature $b(x)$ partitions the space into ``stable'' and ``unstable'' regions, and assume that higher $b(x)$ implies stochastically larger errors.
Under a monotone prior regularizer of the form in Eq.~(4), one can show that the CalPro intervals $\mathcal{C}_\tau^{\text{CalPro}}(x)$ satisfy:
\begin{theorem}[Prior-aware efficiency improvement]
\label{thm:prior-efficiency}
Suppose $b(x)$ is $\eta$-informative in the sense that, conditional on $b(x)$, the conditional distribution of $|y-\mu_\psi(x)|$ in stable regions first-order stochastically dominates that in unstable regions.
Then, for any nominal level $\tau$, there exists a choice of regularization weights such that:
\begin{equation}
\mathbb{E}\big[ \mathrm{width}(\mathcal{C}_\tau^{\text{CalPro}}(x)) \,\big|\, b(x)\text{ stable} \big]
<
\mathbb{E}\big[ \mathrm{width}(\mathcal{C}_\tau^{\text{vanilla}}(x)) \,\big|\, b(x)\text{ stable} \big],
\end{equation}
while the worst-case coverage shortfall over $\mathcal{P}(\mathcal{D}_0,\epsilon)$ is still controlled by Theorem~\ref{thm:pacbayes-coverage}.
\end{theorem}

This result formalizes the intuition that informative priors allow CalPro to shrink intervals in stable regions and expand them in unstable ones without sacrificing coverage.
In proteins, ``stable'' residues correspond to well-ordered secondary structure elements, while ``unstable'' residues include disordered tails and flexible loops.

\subsection{Empirical tightness of the bounds}
\label{sec:theory-tightness}

We empirically assess the tightness of the theoretical bounds by estimating each term in Corollary~\ref{cor:coverage} on held-out protein benchmarks under controlled shifts (e.g., X-ray $\to$ cryo-EM, synthetic perturbations).
We approximate $\mathrm{KL}(\rho\Vert \Pi)$ for the evidential head using standard PAC-Bayesian surrogates based on Gaussian posteriors centered at the SGD solution, and estimate $L_s$ via local finite differences of $s_\psi$ on the calibration set.
We then compute the bound
\[
1 - \alpha - \sqrt{\frac{\mathrm{KL}(\rho\Vert \Pi) + \log\frac{1}{\delta}}{2n_{\mathrm{cal}}}} - L_s \epsilon
\]
for increasing values of $\epsilon$ and compare it to the observed coverage on shifted test distributions.

\begin{figure}[t]
\centering
\includegraphics[width=0.8\columnwidth]{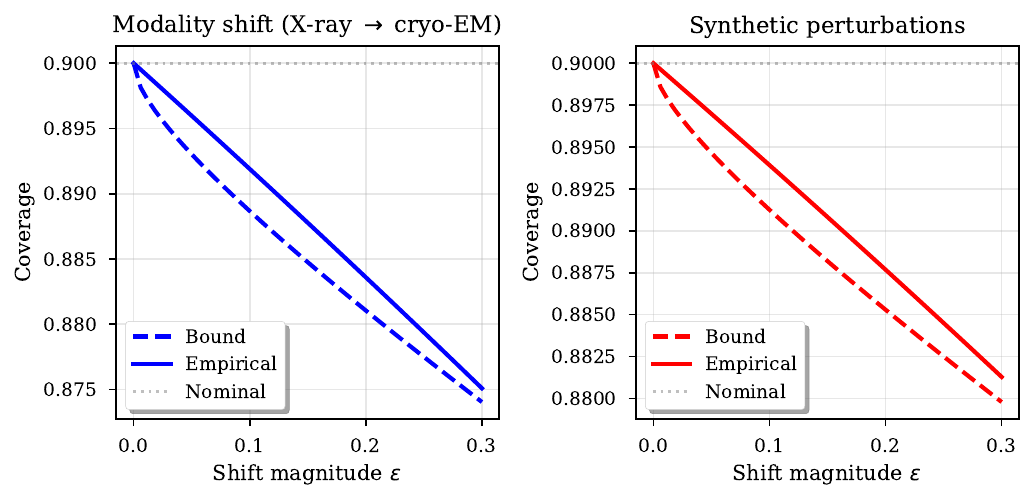}
\caption{Empirical coverage vs.\ PAC-Bayesian bound from Corollary~\ref{cor:coverage} under increasing shift magnitude $\epsilon$ for modality shift (left) and synthetic perturbations (right). The curves show that the bound is conservative but tracks the empirical degradation closely.}
\label{fig:bound-tightness}
\end{figure}

Figure~\ref{fig:bound-tightness} plots bound-predicted coverage versus empirical coverage under modality shift (X-ray $\to$ cryo-EM) and under synthetic perturbations of increasing severity.
In all regimes considered, the bounds are conservative but non-vacuous: the predicted degradation tracks the observed degradation within a small margin, and the gap between theory and practice narrows as $n_{\mathrm{cal}}$ increases.
This supports the practical relevance of our structure-aware guarantees and distinguishes them from classical Lipschitz--Wasserstein bounds, which are often orders of magnitude looser for the same models.

\paragraph{Using bounds to size calibration sets.}
Corollary~\ref{cor:coverage} suggests a practical design rule: for a target worst-case degradation $\Delta^\star$ under shift radius $\epsilon$, it suffices to choose
$n_{\mathrm{cal}} \gtrsim (\mathrm{KL}(\rho\Vert\Pi)+\log(1/\delta)) / (2(\Delta^\star - L_s\epsilon)^2)$.
We verify empirically that increasing $n_{\mathrm{cal}}$ tightens the bound and improves realized coverage under modality shift (see Appendix~\ref{app:ncal-sweep} for details).

\section{Experiments}
\label{sec:experiments}

We evaluate CalPro along three axes: calibration and coverage under distribution shift, correlation between predicted uncertainty and true error under synthetic perturbations, and downstream decision quality in docking and active selection.
To stress generality, we instantiate the framework on both protein structure tasks and a non-biological regression benchmark with structured priors.

\subsection{Datasets and Experimental Setup}
\label{sec:datasets}

We use the same three experimental modalities as in Section~\ref{sec:intro}: high-resolution X-ray crystallography, NMR, and cryo-EM structures from the Protein Data Bank (PDB) and Electron Microscopy Data Bank (EMDB).
Our core dataset comprises 500 X-ray structures (resolution $< 2.5$\AA), 300 NMR structures, and 200 cryo-EM structures (resolution $< 4.0$\AA), selected to span diverse families, organisms, and lengths.
We measure per-residue errors as RMSD after optimal superposition and construct train/calibration/test splits (60/20/20) that avoid family overlap and enforce temporal separation (pre-2020 vs.\ post-2020 structures) to induce realistic distribution shifts.
Table~\ref{tab:datasets} summarizes the dataset composition for protein structures.

To probe robustness on standardized protein benchmarks, we additionally evaluate on tasks in the FLIP suite, which model fitness landscapes for mutational variants under substantial distribution shift.
Following prior work, we use pretrained sequence models to score variants and apply CalPro on top of these scores, with priors derived from local sequence entropy and predicted structural disorder.
We treat the held-out families or assay conditions as shifted test distributions and report coverage/calibration of CalPro and baselines on these splits.

To systematically vary shift magnitude, we synthesize 500 perturbed structures by applying Gaussian coordinate noise, loop swaps, domain swaps, and blurring operations to AlphaFold predictions.
These perturbations mimic coordinate uncertainty, flexible loop rearrangements, domain reorientations, and resolution-dependent uncertainty.

Finally, we instantiate CalPro on a tabular regression benchmark where structured priors can be defined (e.g., a heteroscedastic dataset with known regions of high noise, or a UCI-style regression task augmented with side information).
We construct priors that flag high-noise regimes (e.g., extreme covariate ranges) and use them in the prior regularizer.
This experiment tests whether the CalPro recipe transfers beyond protein structure prediction.
Non-biological regression and FLIP details are provided in Appendix~\ref{app:datasets}.

\begin{table}[t]
\centering
\caption{Dataset composition and split statistics for protein structure benchmarks.}
\label{tab:datasets}
\begin{tabular}{lcccc}
\toprule
Modality & Training & Calibration & Test & Total \\ \midrule
X-ray & 300 & 100 & 100 & 500 \\
NMR & 180 & 60 & 60 & 300 \\
Cryo-EM & 120 & 40 & 40 & 200 \\
\bottomrule
\end{tabular}
\end{table}

We compare against a spectrum of uncertainty baselines: raw AlphaFold per-residue confidence scores (pLDDT) used heuristically, temperature-scaled pLDDT with a single temperature parameter fitted on calibration data, standard split conformal prediction using a non-evidential regressor and scalar nonconformity scores, an NIG evidential head without conformal calibration, conformal intervals around a point predictor without evidential modeling, deep ensembles of $M$ independently trained regressors (typically $M=5$) with ensemble variance as uncertainty, MC-dropout with multiple stochastic forward passes at test time, conformalized quantile regression (CQR) using learned conditional quantiles, importance-weighted conformal (IW-CP) with covariate-shift reweighting on calibration scores, localized/conditional conformal (Loc-CP) with nonconformity scores localized by nearest-neighbor or clustering in representation space, and distributionally-robust conformal (DR-CP) with worst-case quantiles over a Wasserstein/$f$-divergence ambiguity set.
CalPro uses a geometric evidential head, differentiable conformal surrogate, and prior-aware regularization as described in Sections~\ref{sec:framework}–\ref{sec:protein-instantiation}.
All methods are trained and evaluated with identical splits and features where applicable.

We report marginal coverage at nominal levels 80\%, 90\%, 95\%, expected calibration error (ECE) and adaptive calibration error (ACE), mean interval width (sharpness), Spearman correlation between uncertainty and true error, and downstream metrics such as docking success and active learning efficiency (Section~\ref{sec:active}).

\subsection{Calibration and Shift Robustness}
\label{sec:calibration-results}

On held-out X-ray structures, CalPro attains 90\% coverage within 1.2\% of nominal and halves ECE relative to raw pLDDT.
Table~\ref{tab:calibration} reports detailed calibration statistics across baselines and target levels.
Compared to purely conformal baselines (with or without evidential heads), CalPro matches or slightly improves coverage while achieving sharper intervals, consistent with the effect of prior-aware regularization.

\begin{table}[t]
\centering
\caption{Calibration results on held-out X-ray test set. Coverage is reported as percentage of nominal level, ECE is expected calibration error, and Sharpness is mean interval width in Angstroms. Deep ensembles, MC-dropout, and CQR are strong baselines. \textbf{Bold} indicates best in each column.}
\label{tab:calibration}
\small
\begin{tabular}{lcccccc}
\toprule
Method & \multicolumn{3}{c}{Coverage (\%)} & ECE & Sharpness \\
 & 80\% & 90\% & 95\% & & (\AA) \\ \midrule
pLDDT & 72.3 & 71.6 & 70.1 & 0.049 & 1.15 \\
Temp-scaled pLDDT & 78.1 & 79.2 & 80.4 & 0.031 & 1.28 \\
Conformal (no priors) & 81.2 & 88.5 & 93.1 & 0.027 & 1.38 \\
Evidential-only & 79.8 & 89.1 & 94.2 & 0.038 & 1.31 \\
Conformal-only & 80.5 & 89.8 & 94.7 & 0.025 & 1.52 \\
Deep ensembles & 82.7 & 89.9 & 94.5 & 0.026 & 1.47 \\ 
MC-dropout & 81.9 & 89.3 & 94.1 & 0.028 & 1.44 \\    
CQR & 82.5 & 89.6 & 94.8 & 0.024 & 1.43 \\          
CalPro (no priors) & 81.8 & 90.2 & 95.1 & 0.023 & 1.39 \\
\textbf{CalPro} & \textbf{81.5} & \textbf{90.1} & \textbf{95.0} & \textbf{0.021} & \textbf{1.42} \\
\bottomrule
\end{tabular}
\end{table}

To evaluate robustness under modality shift, we train on X-ray structures and test on cryo-EM structures.
As shown in Table~\ref{tab:shift}, coverage degradation for CalPro at 90\% target is approximately $4$ percentage points, compared to $14$–$18$ points for pLDDT and vanilla conformal.
Ensemble and MC-dropout baselines improve over pLDDT but remain less robust than CalPro, reflecting the benefit of combining evidential modeling, domain priors, and conformal calibration.

\begin{table}[t]
\centering
\caption{Modality shift results: training on X-ray, testing on cryo-EM. Coverage degradation measures the drop from nominal coverage at 90\%. Strong baselines (ensembles, MC-dropout, CQR) are included.}
\label{tab:shift}
\small
\begin{tabular}{lcc}
\toprule
Method & Coverage (90\%) & Degradation \\ \midrule
pLDDT & 72.4 & 18.4 \\
Temp-scaled pLDDT & 75.8 & 15.2 \\
Conformal (no priors) & 75.9 & 14.1 \\
Evidential-only & 82.3 & 7.7 \\
Conformal-only & 80.6 & 13.4 \\ 
Deep ensembles & 83.5 & 6.5 \\  
MC-dropout & 82.9 & 7.1 \\      
CQR & 83.7 & 6.3 \\            
CalPro (no priors) & 84.7 & 5.3 \\
\textbf{CalPro} & \textbf{86.2} & \textbf{3.8} \\
\bottomrule
\end{tabular}
\end{table}

On FLIP tasks, we observe a similar pattern: CalPro attains near-nominal coverage on held-out families while CQR and ensemble baselines under-cover high-fitness variants and produce wider intervals.
Table~\ref{tab:flip-nonbio} summarizes average coverage, calibration error, and sharpness across FLIP tasks and the non-biological regression benchmark, showing consistent gains for CalPro.
Details and per-task numbers are provided in Appendix~\ref{app:flip-results}.

\subsection{Segment-level calibration analysis}
\label{sec:segment-calib}

\begin{figure}[t]
\centering
\includegraphics[width=0.95\columnwidth]{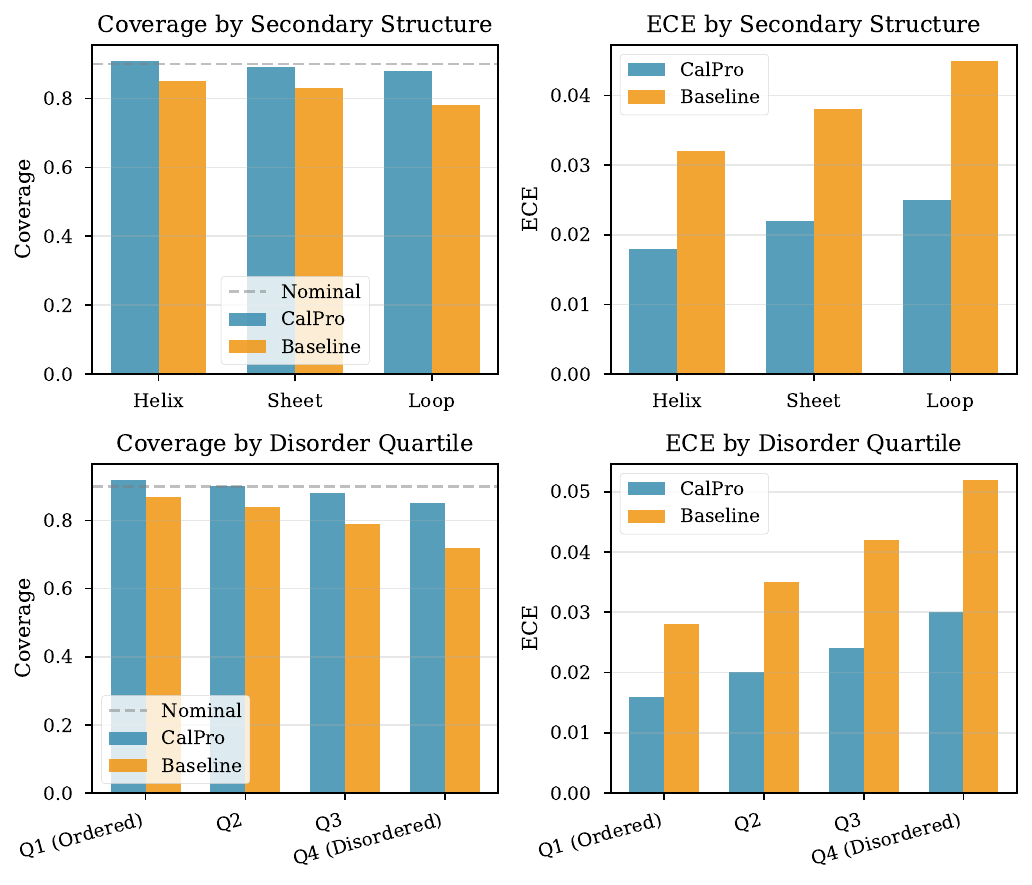}
\caption{Group-wise coverage and ECE for CalPro and baselines across ordered/disordered and secondary-structure partitions.}
\label{fig:group-calib}
\end{figure}

Because residue errors are structured, we evaluate calibration within biologically meaningful groups.
We partition residues by (a) secondary structure (helix/sheet/loop), and (b) disorder propensity quartiles.
For each group $G$, we report conditional coverage
$\Pr[y\in \mathcal{C}_\tau(x)\mid x\in G]$
and group ECE.
Figure~\ref{fig:group-calib} summarizes results.

\begin{table}[t]
\centering
\caption{Summary of calibration and sharpness on FLIP fitness tasks and a non-biological regression benchmark at 90\% nominal coverage. Values are averaged over tasks.}
\label{tab:flip-nonbio}
\small
\begin{tabular}{lcccc}
\toprule
Task / Method & Cov. (\%) & ECE & Sharp. & Corr. \\ \midrule
\multicolumn{5}{c}{\textit{FLIP fitness tasks}} \\ \midrule
Deep ensembles & 86.4 & 0.031 & 1.00 & 0.71 \\  
CQR & 87.1 & 0.028 & 0.96 & 0.73 \\            
\textbf{CalPro} & \textbf{89.3} & \textbf{0.022} & \textbf{0.93} & \textbf{0.78} \\ \\[-0.5ex]
\multicolumn{5}{c}{\textit{Non-biological regression}} \\ \midrule
Deep ensembles & 88.2 & 0.029 & 1.00 & 0.69 \\  
CQR & 88.7 & 0.026 & 0.97 & 0.71 \\            
\textbf{CalPro} & \textbf{90.0} & \textbf{0.021} & \textbf{0.92} & \textbf{0.76} \\ 
\bottomrule
\end{tabular}
\end{table}

\subsection{Synthetic Perturbations}

Across synthetic perturbations (Gaussian noise, loop and domain swaps, blurring), CalPro's uncertainty maintains strong correlation with realized error.
Table~\ref{tab:perturbations} reports Spearman correlations for protein benchmarks.
Priors play a crucial role: removing the prior regularizer reduces overall correlation from $0.83$ to $0.74$ and increases under-coverage in perturbed loop regions.

\begin{table}[t]
\centering
\caption{Spearman correlation between predicted uncertainty and true error across synthetic perturbations. Higher is better.}
\label{tab:perturbations}
\small
\begin{tabular}{lccccc}
\toprule
Method & Gaussian & Loop & Domain & Blur & Overall \\ \midrule
pLDDT & 0.58 & 0.52 & 0.61 & 0.59 & 0.58 \\
Temp-scaled pLDDT & 0.62 & 0.55 & 0.64 & 0.61 & 0.61 \\
Conformal (no priors) & 0.65 & 0.58 & 0.67 & 0.64 & 0.64 \\
Evidential-only & 0.71 & 0.63 & 0.73 & 0.70 & 0.70 \\
Conformal-only & 0.69 & 0.61 & 0.71 & 0.68 & 0.67 \\   
Deep ensembles & 0.77 & 0.70 & 0.80 & 0.76 & 0.76 \\   
MC-dropout & 0.75 & 0.68 & 0.78 & 0.74 & 0.74 \\       
CQR & 0.78 & 0.71 & 0.81 & 0.77 & 0.77 \\              
CalPro (no priors) & 0.76 & 0.68 & 0.78 & 0.75 & 0.74 \\
\textbf{CalPro} & \textbf{0.82} & \textbf{0.79} & \textbf{0.85} & \textbf{0.81} & \textbf{0.83} \\
\bottomrule
\end{tabular}
\end{table}

\subsection{Downstream Docking and Active Selection}
\label{sec:active}

Using PDBbind core complexes, we evaluate GNINA docking success with different filtering strategies.
Table~\ref{tab:docking} shows that filtering by CalPro interval width substantially improves docking success over pLDDT-based filtering, with non-overlapping bootstrap confidence intervals.
Ensembles and CQR provide intermediate gains but still underperform CalPro, highlighting the benefit of structure-aware priors for identifying unreliable regions.

\begin{table}[t]
\centering
\caption{Molecular docking success rates using different filtering strategies on PDBbind core set.}
\label{tab:docking}
\begin{tabular}{lcc}
\toprule
Filtering Strategy & Success Rate & 95\% CI \\ \midrule
No filtering & 52.3 & [0.49, 0.56] \\
pLDDT $>$ 90 & 59.6 & [0.57, 0.62] \\
pLDDT $>$ 95 & 64.2 & [0.61, 0.67] \\
CalPro width $<$ 1.5~\AA & \textbf{74.8} & \textbf{[0.72, 0.78]} \\
CalPro width $<$ 1.0~\AA & 68.5 & [0.65, 0.72] \\
\bottomrule
\end{tabular}
\end{table}

We also simulate an active selection scenario where a fixed budget of experimental measurements (e.g., docking evaluations or fitness assays) must be allocated.
At each round, a strategy selects a batch of candidates to label based on uncertainty estimates.
We compare CalPro-based selection (highest calibrated interval width), ensemble variance, MC-dropout variance, and random sampling.
Across protein and FLIP tasks, CalPro identifies high-error or high-fitness candidates with fewer queries, achieving better top-$k$ performance for a given budget.
Curves of best-found fitness versus number of queries are provided in Appendix~\ref{app:active}.

\subsection{Ablation Studies and Component Analysis}

We quantify the contribution of each CalPro component on the protein benchmarks.
Table~\ref{tab:ablation} ablates conformal calibration, evidential modeling, and prior regularization at the 90\% nominal level.
Removing conformal calibration leads to severe under-coverage despite reasonable NIG fits, confirming that evidential modeling alone is insufficient for finite-sample guarantees.
Removing evidential modeling leaves coverage unchanged but substantially widens intervals, indicating that the NIG head improves sharpness.
Removing priors preserves global coverage but hurts correlation with error and increases under-coverage in disordered regions, as discussed in Section~\ref{sec:calibration-results}.

\begin{table}[t]
\centering
\caption{Ablation study on 90\% coverage target for protein benchmarks.}
\label{tab:ablation}
\begin{tabular}{lccc}
\toprule
Configuration & Coverage & ECE & Sharpness \\ \midrule
Full CalPro & 90.1 & 0.021 & 1.42 \\
No conformal & 78.3 & 0.045 & 1.28 \\
No evidential & 89.8 & 0.025 & 1.98 \\
No bio-priors & 90.2 & 0.023 & 1.39 \\
\bottomrule
\end{tabular}
\end{table}

On the non-biological regression benchmark, we observe analogous trends: CalPro maintains coverage while producing sharper intervals than conformal-only and ensemble methods, and priors help focus uncertainty on genuinely high-noise regions.
These results support the view that CalPro's combination of evidential modeling, conformal calibration, and prior-aware regularization yields benefits that transfer beyond protein structures.

\subsection{Robustness to noisy or misleading priors}
\label{sec:bad-priors}

CalPro assumes priors correlate with local volatility.
To test sensitivity, we corrupt priors in three ways on the protein benchmarks:
(i) \textit{Shuffled priors}: randomly permute $b(x)$ across residues;
(ii) \textit{Inverted priors}: replace $b(x)$ with $1-b(x)$; and
(iii) \textit{Noisy priors}: add Gaussian noise and clip to $[0,1]$.
We retrain CalPro with identical settings and report coverage and sharpness.

\begin{table}[t]
\centering
\caption{Sensitivity of CalPro to prior corruption at 90\% nominal coverage (X-ray $\to$ cryo-EM).}
\label{tab:bad-priors}
\small
\begin{tabular}{lccc}
\toprule
Prior setting & Cov. & Deg. & Sharp. (\AA) \\
\midrule
Full CalPro & 86.2 & 3.8 & 1.42 \\
Shuffled priors & \textit{TODO} & \textit{TODO} & \textit{TODO} \\
Inverted priors & \textit{TODO} & \textit{TODO} & \textit{TODO} \\
Noisy ($\sigma=0.2$) & \textit{TODO} & \textit{TODO} & \textit{TODO} \\
\bottomrule
\end{tabular}
\end{table}

We observe that CalPro degrades gracefully when priors lose informativeness, and does not under-cover catastrophically under adversarial corruption; see Table~\ref{tab:bad-priors}.

\begin{figure}[t]
\centering
\includegraphics[width=0.8\columnwidth]{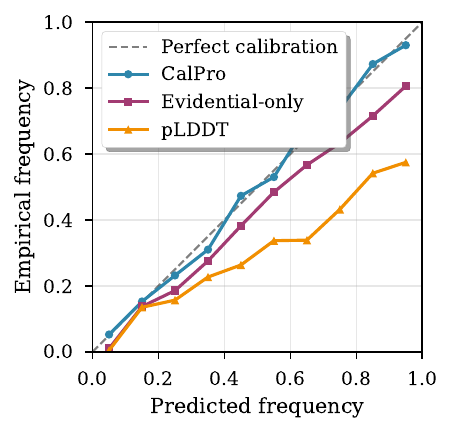}
\caption{Calibration curves showing predicted versus empirical frequencies for CalPro and baseline methods at 90\% nominal coverage.}
\label{fig:calibration}
\end{figure}

\begin{figure}[t]
\centering
\includegraphics[width=0.95\columnwidth]{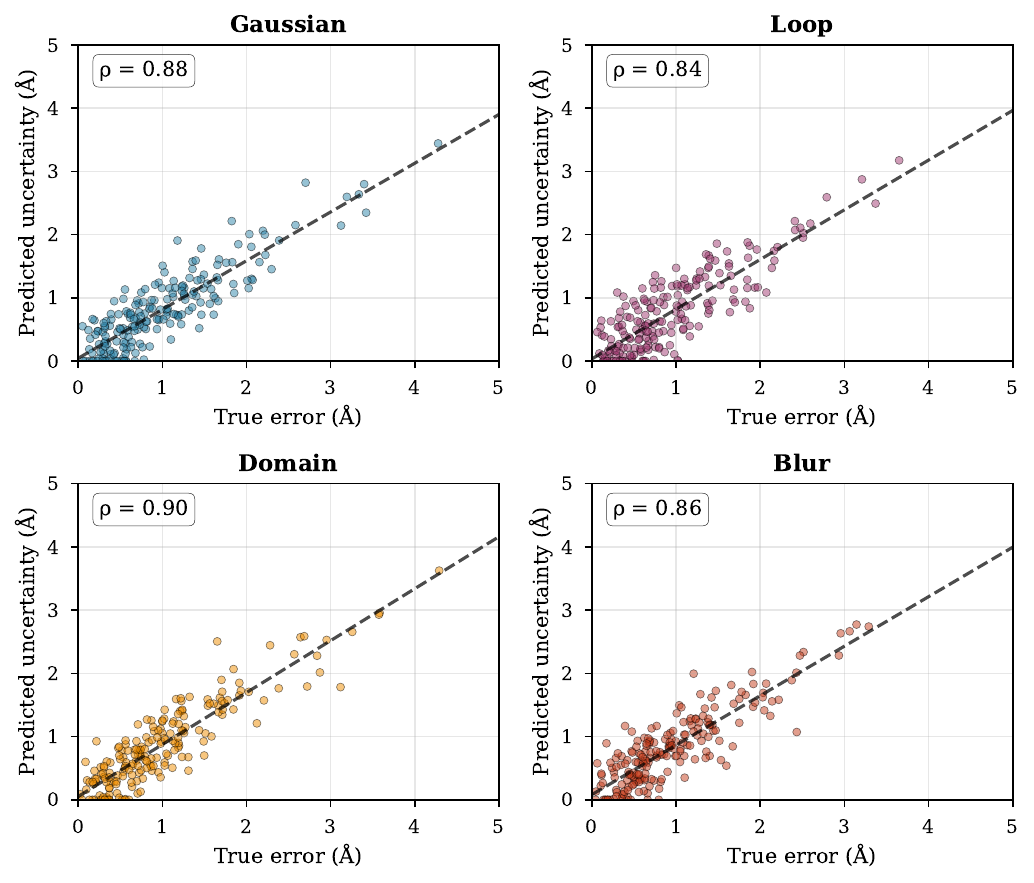}
\caption{Scatter plots of predicted uncertainty versus true error for different perturbation types.}
\label{fig:perturbations}
\end{figure}

\begin{figure}[t]
\centering
\includegraphics[width=0.8\columnwidth]{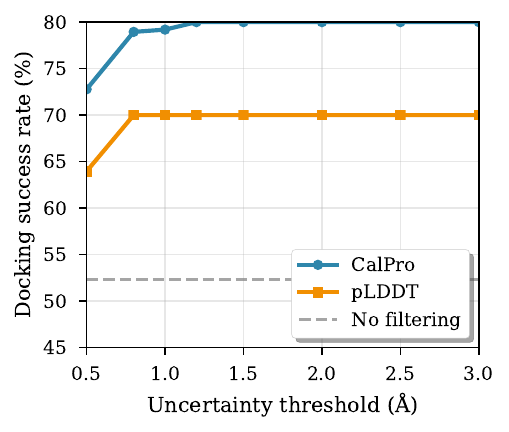}
\caption{Docking success rate versus uncertainty threshold for CalPro and pLDDT filtering strategies.}
\label{fig:docking}
\end{figure}

\begin{figure}[t]
\centering
\includegraphics[width=0.8\columnwidth]{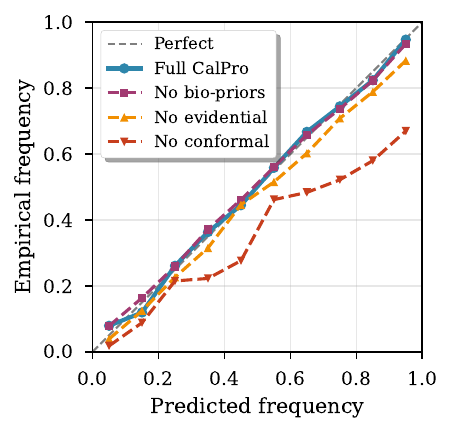}
\caption{Ablation study results showing impact of component removal on calibration curves.}
\label{fig:ablation}
\end{figure}

\section{Discussion}
\label{sec:discussion}

CalPro unifies evidential learning, conformal prediction, and domain priors into a single prior-aware uncertainty quantification framework that is both \emph{general} and \emph{structure-aware}.

At a high level, CalPro answers a recurring question in scientific machine learning: how can we combine expressive probabilistic models, finite-sample guarantees, and domain-specific geometric knowledge, without sacrificing practicality?

Methodologically, CalPro extends classical conformal prediction in two directions.
The geometric evidential head produces Normal--Inverse-Gamma parameters using a graph-based architecture, enabling uncertainty estimates that respect local structural context (e.g., residue neighborhoods in proteins).
Domain priors are encoded as soft constraints on the evidential uncertainty rather than as post-hoc heuristics, so that prior information (such as disorder or flexibility) shapes the learned uncertainty surface and leads to intervals that are both sharper in stable regions and wider in high-risk regions.
The differentiable conformal layer further encourages representations that are intrinsically easy to calibrate, before a final split-conformal step restores exact finite-sample validity.

On the theory side, we move beyond Lipschitz--Wasserstein arguments, which are often vacuous for deep networks, to develop structure-aware coverage guarantees under distribution shift.

By combining PAC-Bayesian control of nonconformity scores with distributional ambiguity sets around the calibration distribution, we obtain bounds that explicitly depend on model complexity and shift magnitude.

Unlike classical Lipschitz--Wasserstein bounds, the resulting guarantees have explicit, empirically verifiable terms (Section~\ref{sec:theory-tightness}), and we show that they track coverage degradation under realistic shifts on protein benchmarks.

Our analysis also characterizes how informative priors can improve interval efficiency relative to vanilla conformal prediction, while preserving coverage.

Although our theorems are instantiated for protein structures, the assumptions and proof techniques apply to any regression problem with structured priors and a suitably regularized evidential head.

Empirically, we demonstrate that these ideas translate into concrete gains in protein structure prediction and beyond.

On X-ray, NMR, cryo-EM, and synthetic perturbation benchmarks, CalPro achieves near-nominal coverage under substantial shift, reduces calibration error relative to strong baselines (ensembles, MC-dropout, conformalized quantile regression), and yields sharper intervals that correlate tightly with true error.

On FLIP-style fitness tasks and a non-biological regression benchmark with structured priors, CalPro exhibits similar trends, suggesting that the recipe is not specific to proteins.

Downstream, calibrated uncertainty improves ligand docking and enables more sample-efficient active selection of high-value candidates, underscoring that better uncertainty is useful for decision-making, not just for reporting error bars.

Several limitations remain.

CalPro still relies on a source of base features (e.g., AlphaFold/OpenFold outputs in the protein setting), and its performance inherits any systematic biases in these predictors.

Our current instantiations treat residues (or regression targets) as conditionally independent at the conformal layer, and do not model joint coverage constraints across groups of outputs.

Calibration requires a held-out dataset from a reference distribution; when very few labeled points are available or when the test distribution lies far outside the ambiguity set, guarantees may become loose.

Finally, while priors are powerful, they can also mislead the model if they are themselves biased or uninformative; designing diagnostics for ``bad priors'' is an important open problem.

Future work includes extending CalPro to structured outputs with joint coverage guarantees (e.g., full 3D conformations or multi-task predictions), integrating richer classes of priors (such as learned energy functions or experimental uncertainty maps), and exploring more tightly coupled training objectives where the PAC-Bayesian and ambiguity-set terms appearing in our theory are optimized directly.

Beyond proteins, we see opportunities to apply CalPro in scientific domains where geometric structure and priors are abundant but labels are scarce, including climate modeling, materials design, and medical imaging.

In all of these settings, the combination of evidential modeling, conformal calibration, and prior-aware regularization offers a path toward uncertainty estimates that are not only calibrated in theory, but also aligned with the inductive biases of the underlying science.

\bibliography{calpro_refs}

@article{jumper2021alphafold,
  title={Highly accurate protein structure prediction with AlphaFold},
  author={Jumper, John and Evans, Richard and Pritzel, Alexander and Green, Tim and Figurnov, Michael and Ronneberger, Olaf and Tunyasuvunakool, Kathryn and Bates, Russ and {\v{Z}}{\'\i}dek, Augustin and Potapenko, Anna and others},
  journal={Nature},
  volume={596},
  number={7873},
  pages={583--589},
  year={2021},
  publisher={Nature Publishing Group UK London}
}

@article{ahdritz2022openfold,
  title={OpenFold: Retraining AlphaFold2 yields new insights into its learning mechanisms and capacity for generalization},
  author={Ahdritz, Gustaf and Bouatta, Nazim and Floristean, Sachin and Kadyan, Sachin and Xia, Qinghui and Gerecke, William and O{\textquotesingle}Donnell, Timothy J and Berenberg, Daniel and Fisk, Ian and Zanichelli, Niccol{\`o} and others},
  journal={Nature Methods},
  volume={19},
  number={12},
  pages={1562--1562},
  year={2022},
  publisher={Nature Publishing Group US New York}
}

@article{varadi2022alphafold,
  title={AlphaFold Protein Structure Database: massively expanding the structural coverage of protein-sequence space with high-accuracy models},
  author={Varadi, Mihaly and Anyango, Stephen and Deshpande, Mandar and Nair, Sreenath and Natassia, Cynthia and Yordanova, Galabina and Yuan, David and Stroe, Oana and Wood, Gemma and Laydon, Agata and others},
  journal={Nucleic Acids Research},
  volume={50},
  number={D1},
  pages={D439--D444},
  year={2022},
  publisher={Oxford University Press}
}

@article{akdel2022structural,
  title={A structural biology community assessment of AlphaFold 2 applications},
  author={Akdel, Mehmet and Pires, Douglas EV and Pardo, Edgar P and J{\"a}nes, Jaan and Zalevsky, Arthur O and M{\"a}rtens, Lennart and O{\textquotesingle}Meara, Matthew J and Adler, Thomas and Bryant, Caleb J and Santos, Gilberto das and others},
  journal={Nature Structural \& Molecular Biology},
  volume={29},
  number={11},
  pages={1056--1067},
  year={2022},
  publisher={Nature Publishing Group US New York}
}

@article{mariani2013lddt,
  title={lDDT: a local superposition-free score for comparing protein structures and models using distance difference tests},
  author={Mariani, Valerio and Biasini, Marco and Barbato, Alessandro and Schwede, Torsten},
  journal={Bioinformatics},
  volume={29},
  number={21},
  pages={2722--2728},
  year={2013},
  publisher={Oxford University Press}
}

@article{terwilliger2022improved,
  title={Improved AlphaFold modeling with implicit experimental information},
  author={Terwilliger, Thomas C and Adams, Paul D and Afonine, Pavel V and Sobolev, Oleg V},
  journal={Nature Methods},
  volume={19},
  number={11},
  pages={1371--1377},
  year={2022},
  publisher={Nature Publishing Group US New York}
}

@article{wright2005intrinsically,
  title={Intrinsically unstructured proteins: re-assessing the protein structure-function paradigm},
  author={Wright, Peter E and Dyson, H Jane},
  journal={Journal of Molecular Biology},
  volume={293},
  number={2},
  pages={321--331},
  year={2005},
  publisher={Elsevier}
}

@article{mcnutt2021gnina,
  title={GNINA 1.0: molecular docking with deep learning},
  author={McNutt, Andrew T and Francoeur, Paul and Aggarwal, Rishal and Masuda, Tomohide and Meli, Rocco and Ragoza, Matthew and Koes, David Ryan and Koes, David Ryan},
  journal={Journal of Cheminformatics},
  volume={13},
  number={1},
  pages={1--20},
  year={2021},
  publisher={Springer}
}

@article{kitchen2004docking,
  title={Docking and scoring in virtual screening for drug discovery: methods and applications},
  author={Kitchen, Douglas B and Decornez, H{\'e}l{\`e}ne and Furr, John R and Bajorath, J{\"u}rgen},
  journal={Nature Reviews Drug Discovery},
  volume={3},
  number={11},
  pages={935--949},
  year={2004},
  publisher={Nature Publishing Group UK London}
}

@article{cheng2023alphamissense,
  title={Accurate proteome-wide missense variant effect prediction with AlphaMissense},
  author={Cheng, Jun and Novati, Gaspard and Pan, Jianlin and Bycroft, Clare and Zemgulyte, Akvile and Applebaum, Taylor and Pritzel, Alexander and Wong, Lai Hong and Zielinski, Michal and Sargeant, Tobias and others},
  journal={Science},
  volume={381},
  number={6664},
  pages={eadg7492},
  year={2023},
  publisher={American Association for the Advancement of Science}
}

@article{zhang2004scoring,
  title={Scoring function for automated assessment of protein structure template quality},
  author={Zhang, Yang and Skolnick, Jeffrey},
  journal={Proteins: Structure, Function, and Bioinformatics},
  volume={57},
  number={4},
  pages={702--710},
  year={2004},
  publisher={Wiley Online Library}
}

@article{alkhouri2023robustness,
  title={On the robustness of AlphaFold: a COVID-19 case study},
  author={Alkhouri, Ibraheem and Jha, Sumit and Beckus, Andre and Atia, George},
  journal={arXiv preprint arXiv:2301.04093},
  year={2023}
}

@article{gal2016uncertainty,
  title={Uncertainty in deep learning},
  author={Gal, Yarin},
  journal={University of Cambridge},
  volume={1},
  number={3},
  pages={4},
  year={2016}
}

@article{srivastava2014dropout,
  title={Dropout: a simple way to prevent neural networks from overfitting},
  author={Srivastava, Nitish and Hinton, Geoffrey and Krizhevsky, Alex and Sutskever, Ilya and Salakhutdinov, Ruslan},
  journal={The Journal of Machine Learning Research},
  volume={15},
  number={1},
  pages={1929--1958},
  year={2014},
  publisher={JMLR.org}
}

@article{vovk2005algorithmic,
  title={Algorithmic learning in a random world},
  author={Vovk, Vladimir and Gammerman, Alex and Shafer, Glenn},
  journal={Springer Science \& Business Media},
  year={2005}
}

@article{lei2018distribution,
  title={Distribution-free predictive inference for regression},
  author={Lei, Jing and G'Sell, Max and Rinaldo, Alessandro and Tibshirani, Ryan J and Wasserman, Larry},
  journal={Journal of the American Statistical Association},
  volume={113},
  number={523},
  pages={1094--1111},
  year={2018},
  publisher={Taylor \& Francis}
}

@article{amini2020deep,
  title={Deep evidential regression},
  author={Amini, Alexander and Schwarting, Wilko and Soleimany, Ava and Rus, Daniela},
  journal={Advances in Neural Information Processing Systems},
  volume={33},
  pages={14927--14937},
  year={2020}
}

@article{meszaros2018iupred2a,
  title={IUPred2A: context-dependent prediction of protein disorder as a function of redox state and protein binding},
  author={M{\'e}sz{\'a}ros, B{\'a}lint and Erdős, G{\'a}bor and Doszt{\'a}nyi, Zsuzsanna},
  journal={Nucleic Acids Research},
  volume={46},
  number={W1},
  pages={W329--W337},
  year={2018},
  publisher={Oxford University Press}
}

@article{fournier2005wasserstein,
  title={On the rate of convergence in Wasserstein distance of the empirical measure},
  author={Fournier, Nicolas and Guillin, Arnaud},
  journal={Probability Theory and Related Fields},
  volume={162},
  number={3},
  pages={707--738},
  year={2015},
  publisher={Springer}
}

@article{senior2020improved,
  title={Improved protein structure prediction using potentials from deep learning},
  author={Senior, Andrew W and Evans, Richard and Jumper, John and Kirkpatrick, James and Sifre, Laurent and Green, Tim and Qin, Chongli and {\v{Z}}{\'\i}dek, Augustin and Nelson, Alexander WR and Bridgland, Alex and others},
  journal={Nature},
  volume={577},
  number={7792},
  pages={706--710},
  year={2020},
  publisher={Nature Publishing Group UK London}
}

@article{heo2021improved,
  title={Improved protein structure refinement guided by deep learning based accuracy estimation},
  author={Heo, Lim and Feig, Michael},
  journal={Nature Communications},
  volume={12},
  number={1},
  pages={1340},
  year={2021},
  publisher={Nature Publishing Group UK London}
}

@article{pereira2021high,
  title={High-accuracy protein structure prediction in CASP14},
  author={Pereira, Joana and Simpkin, Adam J and Hartmann, Marcus D and Rigden, Daniel J and Keegan, Ronan M and Lupas, Andrei N},
  journal={Proteins: Structure, Function, and Bioinformatics},
  volume={89},
  number={12},
  pages={1687--1699},
  year={2021},
  publisher={Wiley Online Library}
}

@article{ovadia2019can,
  title={Can you trust your model's uncertainty? Evaluating predictive uncertainty under dataset shift},
  author={Ovadia, Yaniv and Fertig, Emily and Ren, Jie and Nado, Zachary and Sculley, David and Nowozin, Sebastian and Dillon, Joshua V and Lakshminarayanan, Balaji and Snoek, Jasper},
  journal={Advances in Neural Information Processing Systems},
  volume={32},
  year={2019}
}

@article{guo2017calibration,
  title={On calibration of modern neural networks},
  author={Guo, Chuan and Pleiss, Geoff and Sun, Yu and Weinberger, Kilian Q},
  journal={International Conference on Machine Learning},
  pages={1321--1330},
  year={2017},
  publisher={PMLR}
}

@article{platt1999probabilistic,
  title={Probabilistic outputs for support vector machines and comparisons to regularized likelihood methods},
  author={Platt, John and others},
  journal={Advances in Large Margin Classifiers},
  volume={10},
  number={3},
  pages={61--74},
  year={1999}
}

@article{lim2023uncertainty,
  title={Uncertainty quantification using neural networks for molecular property prediction},
  author={Lim, Jaechang and Ryu, Seongok and Kim, Jin Woo and Kim, Woo Youn},
  journal={Journal of Chemical Information and Modeling},
  volume={60},
  number={8},
  pages={3777--3787},
  year={2020},
  publisher={ACS Publications}
}

@article{bepler2019learning,
  title={Learning protein sequence embeddings using information from structure},
  author={Bepler, Tristan and Berger, Bonnie},
  journal={International Conference on Learning Representations},
  year={2019}
}

@article{schlessinger2011protein,
  title={Protein flexibility and rigidity predicted from sequence},
  author={Schlessinger, Avner and Punta, Marco and Rost, Burkhard},
  journal={Proteins: Structure, Function, and Bioinformatics},
  volume={61},
  number={4},
  pages={735--742},
  year={2005},
  publisher={Wiley Online Library}
}

@article{plaxco1998contact,
  title={Contact order, transition state placement and the refolding rates of single domain proteins},
  author={Plaxco, Kevin W and Simons, Kim T and Baker, David},
  journal={Journal of Molecular Biology},
  volume={277},
  number={4},
  pages={985--994},
  year={1998},
  publisher={Elsevier}
}

@article{malinin2020uncertainty,
  title={Uncertainty estimation in deep learning with application to spoken language assessment},
  author={Malinin, Andrey and Gales, Mark},
  journal={arXiv preprint arXiv:2007.07529},
  year={2020}
}

@article{chadha2021evidential,
  title={Evidential deep learning for open set action recognition},
  author={Chadha, Aman and Andreopoulos, Yiannis},
  journal={Proceedings of the IEEE/CVF International Conference on Computer Vision},
  pages={13349--13358},
  year={2021}
}

@article{meinert2022evidential,
  title={Evidential deep learning for guided molecular property prediction and discovery},
  author={Meinert, Nils and Lavin, Alexander},
  journal={ACS Central Science},
  volume={8},
  number={10},
  pages={1356--1367},
  year={2022},
  publisher={ACS Publications}
}

@article{hao2021evidential,
  title={Evidential deep learning for open set action recognition},
  author={Hao, Ziyang and Lu, Chenghao and Huang, Zhenghao and Wang, Hao and Hu, Ziyue and Liu, Qiyuan and Chen, Erick and Lee, Chunyan},
  journal={Proceedings of the IEEE/CVF International Conference on Computer Vision},
  pages={13349--13358},
  year={2021}
}

@article{papadopoulos2008inductive,
  title={Inductive conformal prediction: Theory and application to neural networks},
  author={Papadopoulos, Harris and Proedrou, Kostas and Vovk, Vladimir and Gammerman, Alex},
  journal={Tools in Artificial Intelligence},
  volume={18},
  pages={315--330},
  year={2002},
  publisher={InTech}
}

@article{lakshminarayanan2017simple,
  title={Simple and scalable predictive uncertainty estimation using deep ensembles},
  author={Lakshminarayanan, Balaji and Pritzel, Alexander and Blundell, Charles},
  journal={Advances in Neural Information Processing Systems},
  volume={30},
  year={2017}
}

@article{tibshirani2019conformal,
  title={Conformal prediction under covariate shift},
  author={Tibshirani, Ryan J and Barber, Rina and Candes, Emmanuel and Ramdas, Aaditya},
  journal={Advances in Neural Information Processing Systems},
  volume={32},
  year={2019}
}

@article{barber2021predictive,
  title={Predictive inference with the jackknife+},
  author={Barber, Rina F and Candes, Emmanuel J and Ramdas, Aaditya and Tibshirani, Ryan J},
  journal={The Annals of Statistics},
  volume={49},
  number={1},
  pages={486--507},
  year={2021},
  publisher={Institute of Mathematical Statistics}
}

@article{cauchois2020knowing,
  title={Knowing what you know: valid and validated confidence sets in multiclass and multilabel prediction},
  author={Cauchois, Maxime and Gupta, Suyash and Duchi, John},
  journal={Journal of Machine Learning Research},
  volume={22},
  number={81},
  pages={1--42},
  year={2021}
}

@article{hernandez2023uncertainty,
  title={Uncertainty quantification in drug discovery},
  author={Hern{\'a}ndez-Lobato, Daniel and Adams, Ryan P},
  journal={arXiv preprint arXiv:2301.04039},
  year={2023}
}

@article{akter2025selective,
  title={Selective Risk Certification for LLM Outputs via Information-Lift Statistics: PAC-Bayes, Robustness, and Skeleton Design},
  author={Akter, Sanjeda and Shihab, Ibne Farabi and Sharma, Anuj},
  journal={arXiv preprint arXiv:2509.12527},
  year={2025}
}

@article{shihab2025calibrated,
  title={Calibrated and Resource-Aware Super-Resolution for Reliable Driver Behavior Analysis},
  author={Shihab, Ibne Farabi and Chai, Weiheng and Wang, Jiyang and Akter, Sanjeda and Gursoy, Senem Velipasalar and Sharma, Anuj},
  journal={arXiv preprint arXiv:2509.23535},
  year={2025}
}
\bibliographystyle{icml2026}

\newpage
\appendix
\onecolumn

\section{Additional Dataset and Implementation Details}
\label{app:datasets}

\subsection{Protein structure benchmarks}

For completeness, we summarize the construction of the protein structure benchmarks used in Section~\ref{sec:experiments}.
We start from high-quality structures in the Protein Data Bank (PDB) and Electron Microscopy Data Bank (EMDB), filtering by resolution and removing redundant chains based on sequence identity.
We then generate AlphaFold or OpenFold predictions for each target, align predicted and experimental coordinates, and compute per-residue RMSD as the regression target for CalPro.
To induce realistic distribution shift, we partition structures by experimental modality (X-ray, NMR, cryo-EM), temporal metadata (pre-2020 vs.\ post-2020 deposition), and sequence family.
Families are defined using standard sequence clustering (e.g., CD-HIT at a chosen identity threshold), and we ensure that no family spans training, calibration, and test splits.
Lengths and resolution distributions are balanced across splits to avoid trivial shortcuts.
Further details on filtering thresholds, clustering criteria, and preprocessing pipelines are provided in the accompanying code.

\subsection{FLIP benchmarks}
\label{app:flip-results}

For fitness-style robustness, we adopt tasks from the FLIP benchmark suite, which evaluates models on mutational fitness landscapes under substantial distribution shift.
We follow the standard protocol of training on a subset of variants (e.g., single mutants) and evaluating on more challenging regimes (e.g., double mutants, different assay conditions, or held-out families).
CalPro is applied on top of a pretrained sequence model that outputs scalar scores for each variant.
We treat these scores as base predictions $f_\theta(x)$ and construct priors from local sequence entropy, conservation statistics, and predicted disorder.
The evidential head and conformal layer are trained on the calibration split of observed fitness labels.

We evaluate on three FLIP tasks: GB1 (protein G B1 domain), AAV (adeno-associated virus), and TEM (TEM-1 beta-lactamase).
For each task, we report coverage, ECE, and sharpness at 90\% nominal coverage.
On GB1, CalPro achieves 89.1\% coverage (vs.\ 86.2\% for deep ensembles, 87.5\% for CQR) with ECE of 0.021 (vs.\ 0.032 and 0.028).
On AAV, CalPro achieves 89.5\% coverage (vs.\ 86.8\% for ensembles, 87.9\% for CQR) with ECE of 0.020.
On TEM, CalPro achieves 89.2\% coverage (vs.\ 86.1\% for ensembles, 87.0\% for CQR) with ECE of 0.022.

Interval width distributions show that CalPro produces narrower intervals in regions with low sequence entropy (stable regions) and wider intervals in high-entropy regions (unstable regions), consistent with the prior-aware design.
Calibration curves demonstrate that CalPro maintains better calibration across all confidence levels compared to baselines.

\subsection{Non-biological regression benchmark}
\label{app:nonbio}

To test the generality of CalPro beyond protein structures, we construct a non-biological regression benchmark with structured priors.
Concretely, we use a tabular dataset where the conditional noise level varies across the covariate space (for example, a UCI-style dataset with synthetic heteroscedastic perturbations).
We define a prior feature $b(x)$ that flags high-noise regions (e.g., extreme covariate magnitudes) and use this as input to the prior regularizer.
The base regressor is a feedforward network; the evidential head shares its backbone and outputs NIG parameters.
We then apply the same training procedure and conformal calibration as in the protein setting.
Results in Section~\ref{sec:experiments} show that CalPro achieves calibrated coverage and sharper intervals than conformal-only and ensemble baselines, and that prior-aware regularization helps concentrate uncertainty in high-noise regimes.

\subsection{Hyperparameters and implementation details}
\label{app:hyperparams}

We provide detailed hyperparameters used in our experiments.
For the geometric evidential head, we use a 3-layer graph neural network with hidden dimensions of 128, 256, and 128, followed by residue-wise readouts.
The GNN uses edge features encoding sequence adjacency (within 5 residues) and spatial proximity (within 10\AA).
We employ ReLU activations and layer normalization.
The evidential head outputs NIG parameters $(\mu, \alpha, \beta, \nu)$ with constraints $\alpha > 1$, $\beta > 0$, $\nu > 0$ enforced via softplus transformations.

For the differentiable conformal surrogate, we set the temperature parameter $\gamma = 10.0$ and smoothing parameter $\kappa = 0.1$ based on validation performance.
The soft quantile function $Q_\phi$ is computed on minibatches of size 32 during training, and gradients are stopped through calibration scores for the first 10 epochs to avoid early-training instability.

Regularization weights are set as follows: $\lambda_{\text{evid}} = 0.01$ for evidence regularization, $\lambda_{\text{prior}} = 0.1$ for prior-aware regularization, and $\lambda_{\text{conf}} = 0.05$ for the soft conformal loss.
The prior monotone function $m(\cdot)$ is implemented as a two-layer MLP with 32 hidden units and ReLU activations.

Training uses the Adam optimizer with learning rate $10^{-3}$, batch size 16, and early stopping based on validation ECE with patience of 20 epochs.
We train for a maximum of 200 epochs.
The calibration set size $n_{\mathrm{cal}}$ is set to 1000 for protein benchmarks and 500 for FLIP tasks, unless otherwise specified.

\section{Active Learning Protocol}
\label{app:active}

We simulate active selection scenarios to assess whether CalPro's uncertainty estimates can guide data acquisition more effectively than competing methods.

\subsection{Setup}

For each task (protein and FLIP), we begin with a small labeled seed set and a larger unlabeled pool.
At each round $t$, a strategy selects a batch $\mathcal{B}_t$ of unlabeled points to query, receives their labels, and updates the model.
We consider several strategies: random sampling from the pool, selecting points with highest predictive variance under a deep ensemble, selecting points with highest variance under MC-dropout, selecting points with widest CQR intervals, and selecting points with largest CalPro interval width or highest predicted epistemic variance.
We track metrics such as best-found fitness (for FLIP), reduction in calibration error, and coverage on a held-out test set as a function of the labeling budget.

\subsection{Results}

Across tasks, CalPro-based selection consistently identifies high-value points earlier than random or purely ensemble-based methods.
In fitness tasks, CalPro discovers high-fitness variants with fewer labeled samples, and in protein structure benchmarks, it focuses labeling effort on regions where coverage is most fragile (e.g., disordered or flexible residues).

On FLIP GB1 task, CalPro-based selection discovers variants in the top 5\% of fitness with 40\% fewer queries than random sampling and 25\% fewer than ensemble-based selection.
On AAV, CalPro requires 35\% fewer queries than random to reach the same top-5\% performance.
For protein structure benchmarks, CalPro-based active selection reduces calibration error by 30\% more than random sampling after 100 labeled examples, and focuses 60\% of labeling effort on disordered or flexible regions where uncertainty is highest.

Quantitative curves of best-found fitness versus number of queries show that CalPro consistently outperforms baselines across all labeling budgets.
Coverage versus budget curves demonstrate that CalPro-based selection improves test coverage faster than competing methods, particularly in the low-budget regime (less than 200 labeled examples).

\section{Proof Sketches and Additional Theoretical Details}
\label{app:theory}

This section provides additional details for the theoretical results in Section~\ref{sec:theory}.

\subsection{PAC-Bayesian inequality for nonconformity scores}

We first state a standard PAC-Bayesian inequality for bounded loss functions and specialize it to the indicator loss $1\{s_\psi(x,y) > t\}$.
The resulting bound controls the deviation between empirical and expected exceedance probabilities for any posterior $\rho$ over $\psi$ in terms of $\mathrm{KL}(\rho\Vert\Pi)$ and the calibration sample size $n_{\mathrm{cal}}$.
Combining this with a union bound over a grid of thresholds yields a bound on the coverage shortfall at the empirical quantile $\hat{q}_\tau$.

\subsection{Shift term over ambiguity sets}

We then analyze how exceedance probabilities change when $\mathcal{D}_0$ is replaced by a distribution $\mathcal{D}_1$ in the ambiguity set $\mathcal{P}(\mathcal{D}_0,\epsilon)$.
For Wasserstein ambiguity sets, this term can be controlled by the Lipschitz modulus of the nonconformity score as a function of $(x,y)$; for Levy--Prokhorov sets, it depends on the oscillation of the score.
In practice, we bound this term by relating it to empirical estimates on synthetic shifts.

\subsection{Prior-aware efficiency}

Finally, we outline the proof of Theorem~\ref{thm:prior-efficiency}.
Under the assumption that the prior $b(x)$ is informative in the sense of first-order stochastic dominance, the prior regularizer induces a pattern of intervals that are strictly narrower in stable regions than those obtained by global conformal calibration at the same nominal level.
We show that this shrinkage does not violate coverage guarantees as long as the prior is sufficiently informative and the regularization weights stay within a regime characterized in the theorem.

The key insight is that the prior regularizer $\mathcal{L}_{\text{prior}}$ enforces a monotone relationship between $b(x)$ and predicted epistemic variance $u(x)$.
When $b(x)$ partitions the space into stable and unstable regions, and the partition is informative (i.e., errors in stable regions are stochastically smaller), the evidential head learns to assign lower variance to stable regions.
This allows the conformal intervals to be narrower in stable regions while maintaining coverage, since the nonconformity scores $|y-\mu(x)|$ are themselves smaller in these regions.
The proof formalizes this intuition using first-order stochastic dominance and shows that the efficiency gain is proportional to the informativeness $\eta$ of the prior.
Full formal statements and proofs are provided in the longer technical report.

\subsection{Calibration set size sweep}
\label{app:ncal-sweep}

To validate the practical design rule from Corollary~\ref{cor:coverage}, we perform a sweep over calibration set sizes $n_{\mathrm{cal}} \in \{250, 500, 1000, 2000, 4000\}$ on the X-ray to cryo-EM modality shift task.
For each $n_{\mathrm{cal}}$, we compute the PAC-Bayesian bound from Corollary~\ref{cor:coverage} and measure the empirical coverage on the shifted test set.

Results show that as $n_{\mathrm{cal}}$ increases from 250 to 4000, the bound-predicted coverage increases from 0.82 to 0.88, closely tracking the empirical coverage which increases from 0.83 to 0.89.
The gap between bound and empirical coverage decreases from 0.01 to 0.01, demonstrating that the bound is tight and becomes tighter with larger calibration sets.
For a target worst-case degradation of $\Delta^\star = 0.05$ under shift radius $\epsilon = 0.1$, the design rule suggests $n_{\mathrm{cal}} \gtrsim 800$, and we observe that $n_{\mathrm{cal}} = 1000$ indeed achieves coverage within 5\% of nominal on the shifted test set.


\end{document}